\documentclass[conference]{IEEEtran}
\IEEEoverridecommandlockouts
\usepackage{amsmath,amssymb,amsfonts}
\usepackage{algorithmic}
\usepackage{graphicx}
\usepackage{textcomp}
\usepackage{xcolor}

\usepackage[
    sorting=none,
    bibstyle=ieee
]{biblatex} 
\begin{document}

\title{Image Denoising Using Convolutional Autoencoder\\

}

\author{\IEEEauthorblockN{Prashanth Venkataraman}
\IEEEauthorblockA{\textit{School of Computer Science and Engineering} \\
\textit{Vellore Institute of Technology}\\
Vellore, India\\
prashanth.v2019@vitstudent.ac.in}

}

\maketitle

\begin{abstract}
With the inexorable digitalisation of the modern world, every subset in the field of technology goes through major advancements constantly. One such subset is digital images which are ever so popular. Images can not always be as visually pleasing or clear as you would want them to be and are often distorted or obscured with noise. A number of techniques to enhance images have come up as the years passed, all with their own respective pros and cons. In this paper, we look at one such particular technique which accomplishes this task with the help of a neural network model commonly known as an autoencoder. We construct different architectures for the model and compare results in order to decide the one best suited for the task. The characteristics and working of the model are discussed briefly knowing which can help set a path for future research.
\end{abstract}


\section{Introduction}
Image denoising refers to the process of removing distortions otherwise known as “noise” from a given input image in order to produce a clearer image. But what is noise exactly and where does it originate from?

Noise refers to random variations in brightness/intensity or color information present on a digital image. Noise has several modes of occurring which include during acquisition, noise caused by analog-to-digital converter errors, bit errors in transmission,statistical quantum fluctuations, faults in the image sensor and many more\cite{fan2019brief}. This noise can render many images useless unless techniques exist to eradicate it.

Image denoising therefore becomes an important challenge which has still not been perfected. More and more new methods come up everyday, each having particular use cases, but none of these techniques have ever been able to perfectly restore an image to its original form.

This paper implements the use of a fully connected, dense autoencoder as well as a convolutional autoencoder for the task. A convolutional autoencoder is a deep learning model built with neural network layers, convolutional layers in specific. Both models are compared with each other to determine the more accurate one.
The following sections will speak in detail about autoencoders and their workings along with their implementation for this problem statement.

\section{The Autoencoder}
The term autoencoder first came about in the 1980s to address the problem of “backpropagation without a teacher”, which is countered by using the input data as the expected output. Along with hebbian learning rules\cite{hebb2005organization}, autoencoders became rudimentary in the field of unsupervised learning. 

The autoencoder\cite{baldi2012autoencoders} is a type of neural network which learns to encrypt/code a given unlabelled input into a dimensional space which may or may not be of the same order as the input, it generally maps the input into a lower dimensional space(latent space). This encoding is then used to reconstruct the original image.

Throughout this process, the model learns the mapping of different input images to specific points in the latent space by training and during this training, the autoencoder is taught to ignore discrepancies or noise present in the input.

Autoencoders are used in a variety of tasks such as image classification, generative modeling, facial detection and so on.

\subsection{Basic Architecture}

An autoencoder generally consists of 2 parts\cite{Wiki}:
\begin{enumerate}
\item An encoder which maps input into the latent space and
\item A decoder which decodes points on the latent space to reconstruct the input.
\end{enumerate}
Instead of just blatantly duplicating the input, an autoencoder is made to deconstruct and reconstruct the input and hence learns to recognise only important parts of the data. A general schema of an autoencoder can be seen in Fig. 1.

\begin{figure}[htbp]
\centerline{\includegraphics[width=90mm,scale=1]{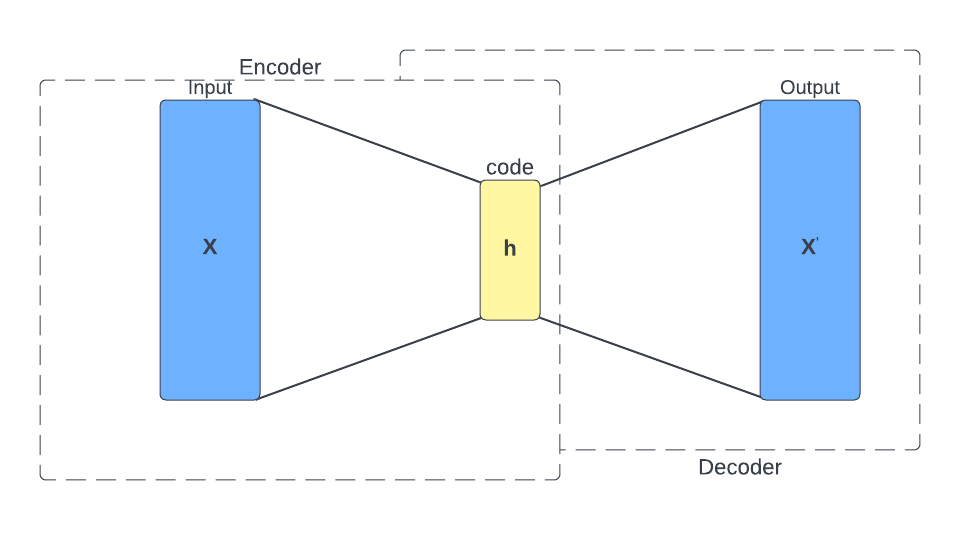}}
\caption{Architecture of a conventional autoencoder}
\label{fig}
\end{figure}

Let the encoder and decoder of the autoencoder be defined as transitions ${\Phi}$ and ${\psi}$ respectively

Given an input $X$

\begin{equation}
  \begin{aligned}
    \Phi: &  X \longrightarrow F\\
    \psi: &  F \longrightarrow X\\
    \Phi, \psi & = argmin_{\Phi,\psi} {||X - (\Phi \circ \psi)X||}^2
  \end{aligned}
\end{equation}

For the case of there being only a single hidden layer, the encoder takes an input $x \in \mathbb{R}^2=X$ and creates a mapping to $h \in \mathbb{R}^p=F$

\begin{equation}
    h=\sigma(Wx+b)
\end{equation}

$h$ is referred to as the coded latent expression of the input, $\sigma$ is a rectified linear unit activation function, $W$ is a weight matrix and b is the bias vector.
The weight matrix and bias vector of a neural network is generally randomly assigned and morph their values via backpropagation during training. 
Finally, the decoder decodes $h$ to a reconstructed version of $x$ denoted as ${x^\prime}$, having the same dimensions as $x$ 

\begin{equation}
    x^\prime=\sigma^\prime(W^\prime h+b^\prime)
\end{equation}

where $\sigma^\prime$ a rectified linear unit activation function and $W^\prime$ and $b^\prime$ are the weight matrix and bias for the decoder and aren't usually related to $W$ and $b$ of the encoder.

The loss of the autoencoder is calculated by how much the input and output differ in terms of individual pixel values. An example is the mean squared error loss.

\begin{equation}
    L(x,x^\prime)=||x-x^\prime||^2 = ||x-\sigma^\prime(W^\prime(\sigma(Wx+b))+b^\prime||^2
\end{equation}
where $x$ is averaged over each training batch.

\subsection{Denoising Autoencoder}
A denoising autoencoder(DAE)\cite{DAE} is quite similar in architecture to a standard autoencoder except that it introduces noise to the input images present in the dataset during training and validation.
The noise introduced is random in nature.

Given a random noise generating function $T$, an input $x$ is given to it which it then converts to a noisy version of $x$ denoted as $T(x)$.
This $T(x)$ is then used as input to the neural network.

\begin{equation}
  \begin{aligned}
    T: &  X \longrightarrow T(X)
  \end{aligned}
\end{equation}

An effective DAE must manage to process any noisy image and reconstruct the original image while eradicating the noise.

A smart way of going about creating the model will involve the use of Convolutional layers in the neural network instead of Dense layers.
This is done because convolutional layers are known to learn really good representations and features of images via their working while full connected layers fail to do so.

\section{model architecture}
The architecture designed for both, a fully connected autoencoder as well as a convolutional autoencoder are described below.
\subsection{fully connected autoencoder}
This model takes in an input of dimension 28*28*1 (grayscale image) and flattens it to an array of dimension 784 which now becomes the input.
This input is passed onto a Dense layer consisting of 64 neurons which needless to say, fully connected to the input.
The output of this Dense layer is passed onto another Dense layer consisting of 784 neurons. The number 784 is chosen so as to reshape the output back into a 28*28 pixel image.
The output of the second Dense layer is reshaped back into a 28*28*1 dimensional array, the same shape as the input, which marks the final output of the encoder.

The architecture can be visualised easier using Fig. 2.
\begin{figure}[htbp]
\centerline{\includegraphics[width=90mm,scale=1]{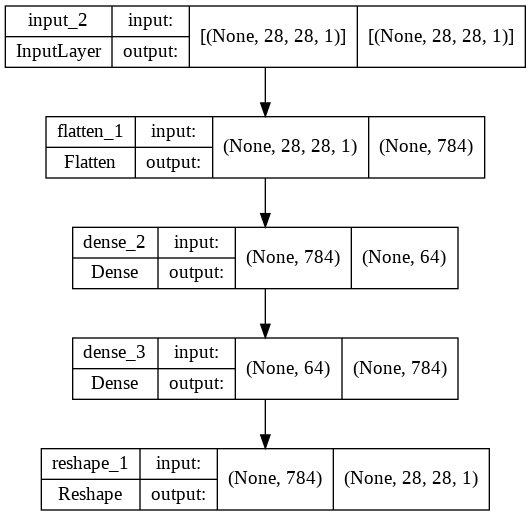}}
\caption{Blueprint of the fully connected autoencoder}
\label{fig}
\end{figure}

\subsection{Convolutional Autoencoder}
A convolutional autoencoder employs the same principle as that of a conventional autoencoder with the only difference being that convolutional layers are used instead of Dense layers. The architecture is inspired from \cite{zhang2018better}.

An input of dimension 28*28*1 is passed through 2 sets of 2D convolutional layers(32 filters, 3*3 weight matrix, relu activation) and encode the input into a latent space of dimension 7*7*32. 
The latent coordinates are then used as input for Conv2D Transpose layers(32 filters, 3*3 weight matrix, relu activation) in order to upscale the image back into its original shape. 

The details of each layer are present in Table 1

\begin{table}[htbp]
\caption{Convolutional autoencoder layer details}
\begin{center}
\begin{tabular}{ | c | c | c | c | c | } 
  \hline
  Layer & Type & Kernel size & Stride & Padding \\
  \hline
  1 & Conv2D & 3 & 1 & same \\ 
  \hline
  2 & Pool(max) & 2 & 1 & same \\
  \hline
  3 & Conv2D & 3 & 1 & same \\
  \hline
  4 & Pool(max) & 2 & 1 & same \\
  \hline
  5 & TransConv2D & 2 & 2 & same \\
  \hline  
  6 & TransConv2D & 2 & 2 & same \\
  \hline
  7 & Conv2D & 3 & 1 & same \\
  \hline
\end{tabular}
\end{center}
\end{table}

Detailed architecture of this network is depicted in Fig. 3.
\begin{figure}[htbp]
\centerline{\includegraphics[width=90mm,scale=1]{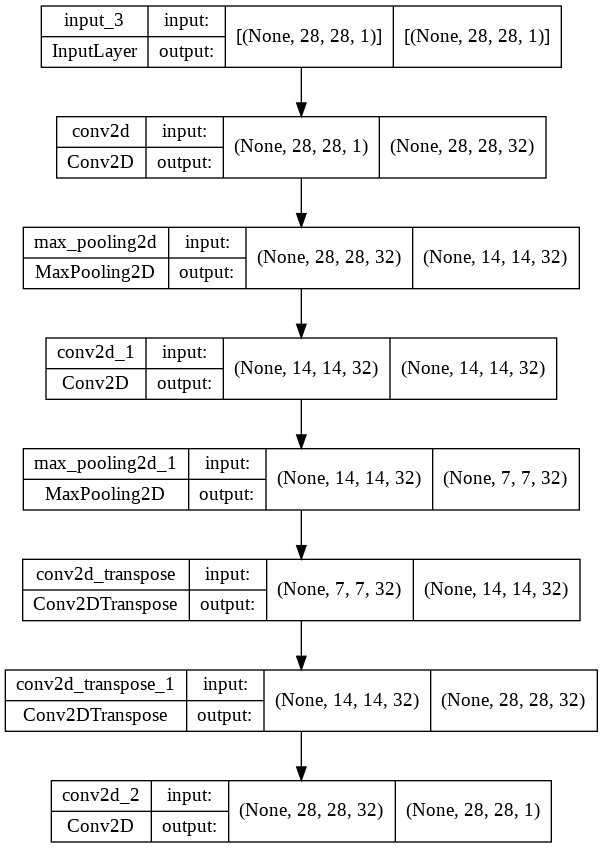}}
\caption{Blueprint of the convolutional autoencoder}
\label{fig}
\end{figure}

Table 2 shows the total number of trainable parameters for each model.
\begin{table}[htbp]
\caption{Trainable parameters}
\begin{center}
\begin{tabular}{ | c | c | } 
  \hline
  Model & Trainable parameters \\
  \hline
  Fully connected autoencoder & 101,200 \\ 
  \hline
  Convolutional autoencoder & 28,353 \\ 
  \hline
\end{tabular}
\end{center}
\end{table}

\section{Results and discussion}
\subsection{Dataset}
The dataset used for training and testing is the MNIST handwritten digits(grayscale images) dataset\cite{mnist} consisting of 60,000 examples in the training set and 10,000 examples in the test set each of shape 28*28*1 pixels.
A few examples from the training set are shown in Fig. 4.

\begin{figure}[htbp]
\centerline{\includegraphics[width=90mm,scale=1]{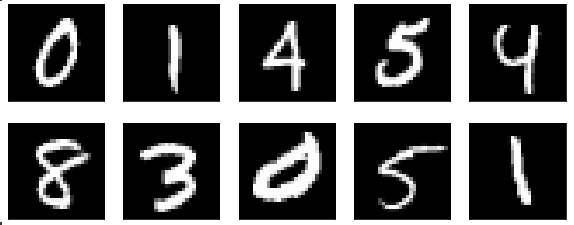}}
\caption{Sample images from the MNIST dataset}
\label{fig}
\end{figure}
Since the dataset we require for the task at hand requires noisy images, we define a functions to add random gaussian noise to the dataset elements. The new dataset now consists of noisy images and can be visualised in Fig. 4.
\begin{figure}[htbp]
\centerline{\includegraphics[width=90mm,scale=1]{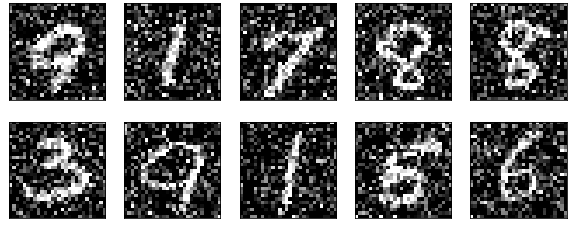}}
\caption{Sample images from the MNIST dataset after the addition of gaussian noise}
\label{fig}
\end{figure}

\subsection{experimental setup}
The neural networks are modelled according to the ones shown in Fig. 2 and Fig. 3 using tensorflow and keras libraries.
The training is done on Google Colab’s Tesla K80 Graphics card having 2496 CUDA cores with 12GB of GDDR5 VRAM.

\subsection{training details}
Each model is trained for a total of 20 epochs and the training loss and validation loss is displayed at the end of each epoch

Both models utilise the Adam optimizer and the loss function used is "binary crossentropy".

Training time for each model amounted to about 100 seconds each due to the input images being of small size and hence not requiring too much processing power.

\subsection{discussions}
The final values for validation loss for each model is depicted in Table 3.

\begin{table}[htbp]
\caption{Model loss}
\begin{center}
\begin{tabular}{ | c | c | } 
  \hline
  Model & Validation loss \\
  \hline
  Fully connected autoencoder & 0.2305\\ 
  \hline
  Convolutional autoencoder & 0.0871 \\ 
  \hline
\end{tabular}
\end{center}
\end{table}

Although the loss of both models don't seem to be that drastically different in value, the predictions of each model tell us a different story.
The predictions made by the fully connected autoencoder is shown in Fig. 6 along with the corresponding inputs.

\begin{figure}[htbp]
\centerline{\includegraphics[width=90mm,scale=1]{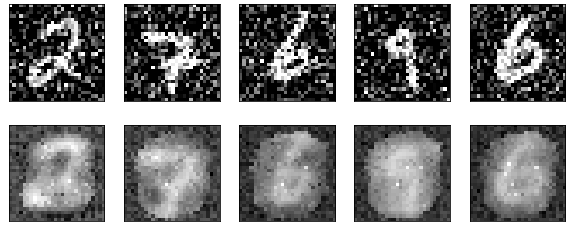}}
\caption{Noisy inputs(top) along with the corresponding denoised predictions(bottom) of the fully connected autoencoder}
\label{fig}
\end{figure}
As it's evident from Fig. 6, the fully connected autoencoder which consists of only Dense layers does not do a good job at denoising, instead it seems to make the input even more noisier.
This is due to the fact that Dense layers don't do a very good job at learning the features and representations for image related tasks.

Whereas, the predictions along with input for the convolutional autoencoder and shown in Fig. 7.

\begin{figure}[htbp]
\centerline{\includegraphics[width=90mm,scale=1]{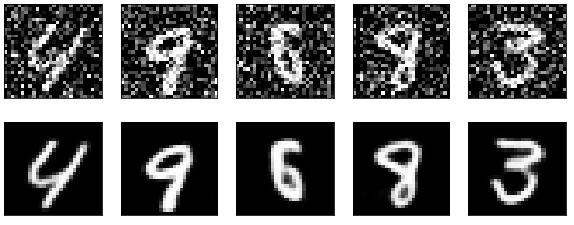}}
\caption{Noisy inputs(top) along with the corresponding denoised predictions(bottom) of the convolutional autoencoder}
\label{fig}
\end{figure}

The difference in the predictions made by the two models in terms of output is vast, it's blatant from Fig. 7 of how much better the convolutional autoencoder is at denoising. Almost all the noise from the input is removed resulting in an almost noise-free, clear image.
This is because of the convolutional layer's incisive ability to store meaningful representations of input during training and hence alienating any noise when mapping the image to a latent space.

\section{conclusion}
The task of image denoising 28*28 pixel, greyscale images using autoencoders is hence visualised. Distinctions between two different autoencoder architectures, namely a Dense autoencoder and a convolutional autoencoder are drawn with which we can ascertain that the use the convolutional layers is very important in the case of image denoising using deep learning. More complex denoising involving RGB images can hence be done by creating a deeper model and more apt latent space dimension. Future research may include creating a model to denoise images of a higher calibre and those which contain multiple channels.

\printbibliography

\end{document}